# A probabilistic approach to hierarchical model-based diagnosis

Sampath Srinivas*
Computer Science Department
Stanford University
Stanford, CA 94305
srinivas@cs.stanford.edu

## Abstract

Model-based diagnosis reasons backwards from a functional schematic of a system to isolate faults given observations of anomalous behavior. We develop a fully probabilistic approach to model based diagnosis and extend it to support hierarchical models. Our scheme translates the functional schematic into a Bayesian network and diagnostic inference takes place in the Bayesian network. A Bayesian network diagnostic inference algorithm is modified to take advantage of the hierarchy to give computational gains.

## 1 INTRODUCTION

Fault diagnosis in engineering systems is a very important problem. The problem is as follows: From observations of anomalous behavior of a system one has to infer what components might be at fault.

Diagnosis fundamentally involves uncertainty. For any reasonable sized system, there is a very large number of possible explanations for anomalous behavior. Instead of reasoning with all of them we want to concentrate on the most likely explanations. In this paper we describe a method for doing model-based diagnosis with a fully coherent probabilistic approach. To do so, we translate the system model into a Bayesian network and perform diagnostic computations within the Bayesian network.

We then extend the notion of system models to include hierarchical models. Hierarchical compositional modeling is an all pervasive technique in engineering practice. It allows modularization of the modeling problem, thus aiding the modeling process. In addition, the hierarchy allows gains in computational tractability. We show how this improvement in tractability extends to diagnosis by describing a hierarchical version of a Bayesian network inference algorithm which takes advantage of the hierarchy in the model to give computational gains.

## 2 THE TRANSLATION SCHEME

In this section we describe how the Bayesian network is created from the system functional schematic. The system functional schematic consists of a set of components. Each component has a set of discrete valued inputs $I_1, I_2, \ldots, I_n$ and a discrete valued output $O$. The component also has a discrete valued *mode* variable $M$. Each state of $M$ is associated with an operating region of the device. Each state of $M$ is associated with a specific input-output behavior of the component.

The component specification requires two pieces of information– a function $F : I_1 \times I_2 \ldots I_n \times M \to O$ and a prior distribution over $M$. The prior distribution quantifies the *a priori* probability that the device functions normally. As an example, a component might have only two possible mode states **broken** and **ok**. If it is very reliable we might have a very high probability assigned to $P(M = \text{ok})$, say 0.995. The components are connected according to the signal flow paths in the device to form the system model (we do not allow feedback paths).

A Bayesian network fragment is created for a component as follows. A node is created for each of the input variables, the mode variable and the output variable. Arcs are added from each of the input variables and the mode to the output variable. The distribution $P(O|I_1, I_2, \ldots, I_n, M)$ is specified by the component function $F$. That is[1], $P(O = o|I_1 = i_1, I_2 = i_2, \ldots, I_n = i_n, M = m) = 1$ iff $F(i_1, i_2, \ldots, i_n, m) = o$. Otherwise the probability is 0. The variable $M$ is assigned the prior distribution

---

*Also with Rockwell International Science Center, Palo Alto Laboratory, Palo Alto, CA 94301.

[1]We use $x$ to denote a state of a discrete variable $X$.

given as part of the component specification.

The network fragments are now interconnected as follows: Whenever the output variable $O^1$ of a component $C^1$ is connected to the input $I_j^2$ of a component $C^2$, an arc is added from the output node $O^1$ of $C^1$ to the input node $I_j^2$ of $C_2$. This arc needs to enforce an equality constraint and so we enter the following distribution into node $I_j^2$: $P(I_j^2 = p|O^1 = q) = 1$ iff $p = q$, otherwise the probability is 0. After interconnecting the Bayesian network fragments created for each component we have a nearly complete Bayesian network. We now make some observations. The network created is indeed a DAG (and hence fulfills one of the necessary conditions for us to claim it is a Bayesian network). This is so because we did not allow any feedback in the original functional schematic.

The probability distribution for every non-root node in the Bayesian network has been specified. This is because every non-root node is either (a) an output node or (b) an input node which is connected to a preceding output node. The probability distribution for every output node has been specified when creating the Bayesian network fragments. The probability distribution for every input node which has an output node as a predecessor has been specified when the fragments were interconnected.

The root nodes in the network fall into two classes. The first class consists of nodes corresponding to mode variables and the second class consists of nodes corresponding to some of the input variables. We note that the marginal probability distributions of all nodes in the first class (i.e, mode variables) have been specified.

The set of variables associated with this second class of nodes are those variables which are inputs to the entire system – that is, these variables are inputs of components which are not downstream of other components. We will call this set of variables *system input* variables. Let us assume that the inputs coming from the environment to the system are all independently distributed. Further let us assume for now that we have access to a marginal distribution for each system input variable[2]. We enter the marginal distribution for each system input variable into its corresponding node. We now have a fully specified Bayesian network.

Consider the original functional schematic. We can interpret every component function and interconnection in the original functional schematic as a constraint on the values that variables in the schematic can take (in the constraint satisfaction sense). We note that the Bayesian network that we have constructed enforces exactly those constraints that are present in the original schematic and no others. Further, it explicitly includes all the information we have about marginal distributions over the mode variables and the system input variables. The Bayesian network is therefore a representation of the joint distribution of the variables in the functional schematic and the mode variables.

We proceed now to use the Bayesian network for diagnosis in the standard manner. Say we make an observation. An observation consists of observing the states of some of the observable variables in the system. As an example, we might have a observation which consists of the values (i.e., states) of all the system input variables and the output values of some of the components. We declare the observation in the Bayesian network. That is, we enter the states of every observed variable into the Bayesian network and then do a belief update with any standard Bayesian network inference algorithm (for example, [Lauritzen88],[Jensen90]).

Say an observation $\Omega = <Y_1 = y_1, Y_2 = y_2, \ldots, Y_k = y_k>$ has been made. After a Bayesian network algorithm performs a belief update we have the posterior distribution $P(X|\Omega)$ available at every node $X$ in the Bayesian network. The posterior distribution on each of the mode variables gives the updated probability of the corresponding component being in each of its modes. This constitutes diagnosis.

We illustrate our scheme with a simple example from the domain of Boolean circuits. The circuit is shown in Fig 1(a). We treat this circuit as our input functional schematic. A particular observation (i.e., input and output values) is marked on the figure. We note that if the circuit was functioning correctly the output for the marked inputs should be 0. Instead the output is a 1. We assume, for this example, that each gate has two possible states for the mode variable, **ok** and **broken**. The modeler provides a prior on the mode of each gate – for each gate the prior probability of it being in the **ok** state is shown next to it in Fig 1(a). We also require a full fault model – i.e., for each gate we should have a fully specified function relating inputs to the output even if the mode of the gate is **broken**. We assume a "stuck-at-0" fault model – i.e., if the gate is in state **broken** the output is 0 irrespective of what the input is. When the gate is in state **ok** the function relating the inputs to the output is the usual Boolean function for the gate.

The Bayesian network corresponding to this schematic is shown in Fig 1(b). We assume that the inputs are independently distributed. We also assume a uniform distribution as the prior for each of the inputs $I_1$, $I_2$ and $I_3$. Note that in this example, any (strictly positive) prior could be assumed without affecting the results of the diagnosis. This is because the state of the input is fully known when the diagnosis is performed. The observation is entered into the network and inference is performed. The posterior probability of being in the **ok** state for each gate is as shown in Fig 1(a).

---

[2]If every observation of the system is guaranteed to contain a full specification of the state of the input, then the actual choice of priors is irrelevant [Srinivas94].

540    Srinivas

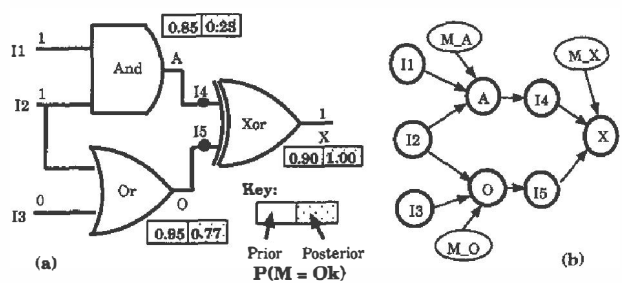

Figure 1: An example: (a) functional schematic (b) corresponding Bayesian network.

## 3  HIERARCHICAL MODELS

Consider a situation where the modeler has conceptually broken up an engineering artifact into a set of component subsystems. She would probably not have a complete functional description (i.e., the function relating inputs to outputs) at this level of abstraction. Each of the component systems has to be modeled at a lower level of detail. We extend our scheme to support such a feature. The modeler first fully specifies the inputs, output and the mode variable of the component. By full specification we mean that the modeler specifies the number of inputs, the possible states of each input variable, the possible states of the output variable and the possible states of the mode variable.

If the modeler would now like to model the component at a lower level of abstraction she can specify a new functional schematic as a detailed description of the component. This new functional schematic would have new components (we will call them subcomponents) which are interconnected to form a functional schematic. This lower level schematic is constrained in the following way: The system input variables of this functional schematic should be the same as the input variables to the component specified at the higher level. Similarly the system output variable of the schematic should be the same as the component output variable at the higher level.

The modeler has to provide a final piece of information to complete the hierarchy – she has to relate the modes of the subcomponents to the modes of the component. To make this more concrete, consider a component which has two states for its mode variable – **ok** and **broken**. Say that it is modeled at a lower level of detail with 4 subcomponents, each of which has two possible states. If we consider the possible combinations of mode states at the lower level of abstraction there are $2^4 = 16$ possibilities. However at the higher level of abstraction there are only two possibilities, i.e., the granularity is not fine enough to distinguish individually between the 16 different possibilities at the lower level.

To relate the lower level to the higher level the modeler has to provide a function describing how the lower level combinations of mode states relate to the higher level mode state. In other words, the modeler has to provide a categorization which separates the lower level state combinations into a set of bins. Each bin corresponds to one of the states of the mode variable at the higher level of abstraction. This function could be a simple rule. One possibility, for example, is the rule "If anything is broken at the lower level then consider the component broken at the higher level". This means, in our example, that 15 possibilities at the lower level fall into the **broken** bin at the higher level while only 1 possibility (i.e., no subcomponents broken) falls into the the **ok** bin at the higher level.

Once this function is specified the hierarchical model is complete. We will call this function the *abstraction function*. Note that we can have multiple levels of hierarchy. We also note two salient points – the modeler does *not* need to provide a component function at higher levels of the hierarchy. In addition the modeler does *not* need to provide a prior on the mode variable at higher levels of hierarchy. In other words, if a component is modeled at a lower level of detail then only the low level functional schematic and the abstraction function are required. The component function and prior are required only for a component which is being modeled "atomically", i.e., it is not being modeled at any finer level of detail.

As an example of hierarchical modeling, consider an exclusive-OR ($XOR$) gate. We might represent the $XOR$ gate at a lower level of detail and show that it is synthesized using $AND$ gates, $OR$ gates and inverters (Fig 2). We use the following rule as the abstraction function: "If anything is broken at the lower level then the $XOR$ gate is broken".

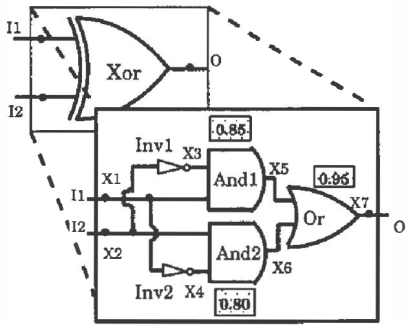

Figure 2: An XOR gate: An example of a hierarchical schematic.

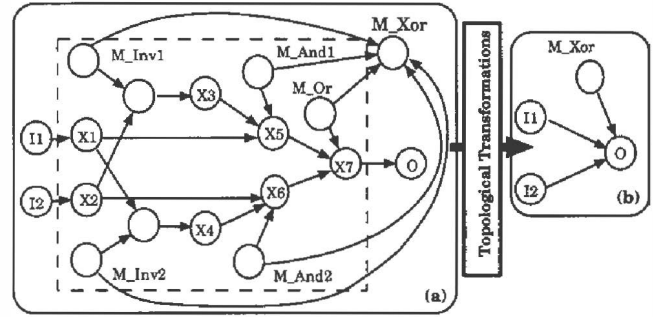

Figure 3: (a) Bayesian network fragment for XOR schematic (b) The fragment after "compilation".

## 3.1 INCORPORATING HIERARCHY IN THE TRANSLATION

When a component is modeled at a lower level, the translation proceeds as follows: Assume that the higher level abstraction does not exist and just plug in the lower level functional schematic between the system inputs and outputs and do the translation. In the resulting Bayesian network introduce a new variable for the higher level mode. Call this $M^h$. Add an arc from the mode variable of each of the subcomponents to the higher level mode variable. Call the lower level mode variables $M^{l1}, M^{l2}, \ldots, M^{ln}$. Fill out the conditional probability distribution of the higher level mode variable as follows: $P(m^h|m^{l1}, m^{l2}, \ldots m^{ln}) = 1$ iff $m^h = Ab(m^{l1}, m^{l2}, \ldots, m^{ln})$, 0 otherwise. Here $Ab$ is the abstraction function relating combinations of mode states of the subcomponents to the mode of the higher level component. Fig 3(a) shows the Bayesian network for the $XOR$ gate example.

Hierarchical models usually have two major and related advantages. The first advantage is that modeling becomes easier. This is because the system is decomposed in a compositional fashion into components with well defined boundaries and interactions. The second advantage is that computation with the model becomes easier. As a first cut, diagnosis with a hierarchical functional model can proceed exactly as described with non-hierarchical models. If we want a fine grain diagnosis we look at the updated posterior probabilities of the subcomponent modes. If we want a coarse grained diagnosis we look at the updated posterior of the mode variable of the component at the higher level of abstraction. However, this simplistic solution does not get any computational gains from the hierarchy.

To get computational gains we need to be able to reason with the higher level model in a way such that the detail of the lower level model has been "compiled away" into a more succinct higher level model. We now describe a scheme for doing so. Consider a component $C^h$ which is modeled at a lower level of abstraction with a functional schematic consisting of subcomponents $C^{l1}, C^{l2}, \ldots, C^{ln}$. The mode variable of $C^h$ is $M^h$ and the mode variable of subcomponent $C^{li}$ is $M^{li}$. Let the inputs of $C^h$ be $I^h_1, I^h_2, \ldots, I^h_m$. Let the output of $C^h$ be $O^h$. Let all the internal variables of the lower level functional schematic (i.e., the input and output variables of the subcomponents excluding the system inputs and outputs) be $X_1, X_2, \ldots, X_k$.

For simplicity, let us assume that all the inputs of $C^h$ are system inputs – i.e., there are no components upstream of $C^h$. We also assume, as described before, that we have a prior on each system input. Now consider the Bayesian network fragment created by the translation scheme for $C^h$. We note that this fragment happens to be a fully specified Bayesian network.

A Bayesian network is a structured representation of the joint distribution of all the variables in the network. In this case the network is a representation following distribution $P(\ I^h_1, I^h_2, \ldots, I^h_m, O^h, M^h, M^{l1}, M^{l2}, \ldots, M^{ln}, X_1, X_2, \ldots, X_k\ )$. Call this the lower level distribution.

If now, we wanted to have a Bayesian network representation at the higher level of abstraction we would not want to explicitly represent the detail about internal variables of the lower level functional schematic or the mode variables of the subcomponents. In other words we would like to have a Bayesian network which represents the joint distribution only the input, mode and output variables of $C^h$, i.e., the distribution $P(I^h_1, I^h_2, \ldots, I^h_m, O^h, M^h)$. Call this the higher level distribution.

We can generate the higher level distribution from the lower level distribution by simply marginalizing out all the irrelevant variables, viz, $M^{l1}$,

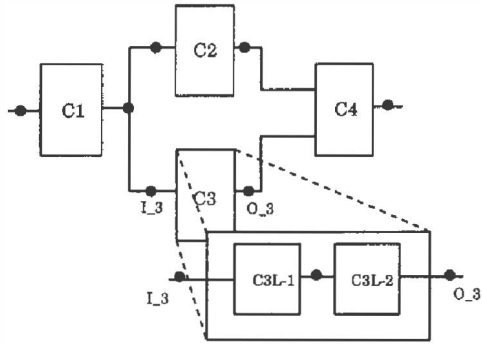

Figure 4: A hierarchical schematic.

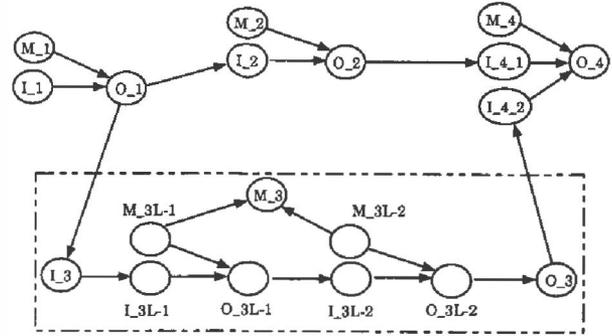

Figure 5: Bayesian network $B^c$ created from schematic of Fig 4.

$M^{l2}, \ldots, M^{ln}, X_1, X_2, \ldots, X_k$. Ideally, we should do this marginalization in some efficient way. Such efficient marginalization is possible using topological transformations of Bayesian networks [Shachter86]. Specifically, we can use the *arc reversal* and *node absorption* operations as follows:

1. Successively reverse the arcs $M^{l1} \rightarrow M^h$, $M^{l2} \rightarrow M^h, \ldots, M^{ln} \rightarrow M^h$. At the end of this step $M^h$ is a root node.

2. Let $\mathbf{X}$ be the set of internal variables of the lower level functional schematic, i.e., $\mathbf{X} = \{M^{l1}, M^{l2}, \ldots, M^{ln}, X_1, X_2, \ldots, X_k\}$. Sort $\mathbf{X}$ into a sequence $\mathbf{X}_{seq}$ in inverse topological order (descendants first). Successively absorb the nodes in $\mathbf{X}_{seq}$ (in order) into $O^h$.

This completes the process and leaves us with the topology shown in Fig 3(b). The successive absorption in the last step is always possible since there is no node $N$ in the Bayesian network such that (a) $N$ is not in $\mathbf{X}_{seq}$ and (b) the position of $N$ has to necessarily be between two nodes contained in $\mathbf{X}_{seq}$ in a global topological order [Shachter86]. Note that the topology which results from the marginalization process described above is the same as the one we would get if we had directly modeled $C^h$ as an atomic component.

For simplicity of exposition, the description above assumes that $C^h$'s inputs are system inputs. However, this assumption is unnecessary. The identical marginalization process is possible for any hierarchically modeled component. We can consider the marginalization process that gives us the higher level distribution as a "compilation process" which is carried out after the model is created.

### 3.2 INTEGRATING HIERARCHY AND DIAGNOSTIC INFERENCE

The hierarchy in the functional schematic can be exploited to improve diagnostic performance. We now describe a method of tailoring the clustering algorithm [Lauritzen88, Jensen90, Pearl88] for Bayesian network inference to take advantage of the hierarchy. This is the most widely used algorithm in practice. The clustering algorithm operates by constructing an tree of cliques from the Bayesian network as a pre-processing step. This construction is by a process called triangulation [Tarjan84]. The resulting tree is called the join tree. Each clique has some of the Bayesian network nodes as its members. As evidence arrives, a distributed update algorithm is applied to the join tree and the results of the update are translated back into updated probabilities for the Bayesian network nodes. The update process mentioned above can be carried out on any join tree that is legal for the Bayesian network.

We will now describe a method of constructing a legal join tree that is tailored to exploit the hierarchy. We explain by means of an example. Consider the hierarchical functional schematic shown in Fig 4. This results in the hierarchical Bayesian network $B^c$ shown in Fig 5.

After the lower level detail is compiled out we get the network $B^h$ in Fig 6(a). We add a dummy node $D^h$ to this Bayesian network such that $M_3$, $I_3$ and $O_3$ are parents of $D^h$. If we run a triangulation algorithm on this network we get a join tree $JT^h$ (Fig 7(a)). We note there exists a clique $\delta^h$ in $JT^h$ such that $I_3$, $M_3$ and $O_3$ belong to $\delta^h$. This is because $I_{3,}$, $M_3$ and $O_3$ are parents of $D^h$. Triangulation guarantees that a Bayesian network node and its parents will occur together in at least one clique in the join tree.

Now consider the lower level network fragment by itself (Fig 6(b)). Call this $B^l$. Say we create a dummy node $D^l$ and add arcs into it from $I_3$, $M_3$ and $O_3$ as shown in the figure. If we triangulate the graph we get a join tree $JT^l$ (Fig 7(b)). Once again, we are guaranteed that there is a clique $\delta^l$ in $JT^l$ such that $I_3$, $M_3$ and $O_3$ belong to $\delta^l$.

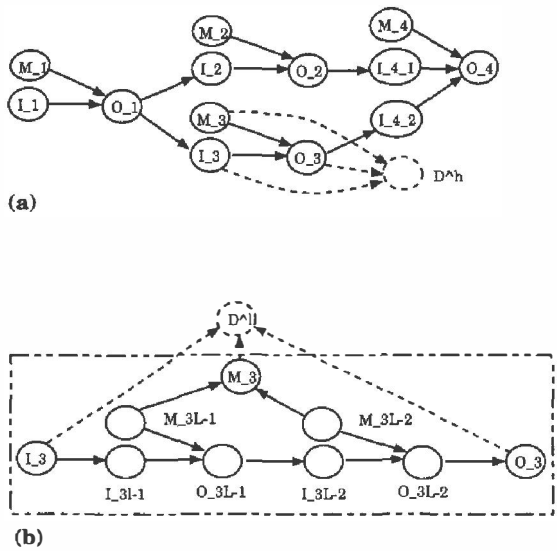

Figure 6: (a) Compiled network $B^h$. (b) Lower level Bayesian network fragment $B^l$.

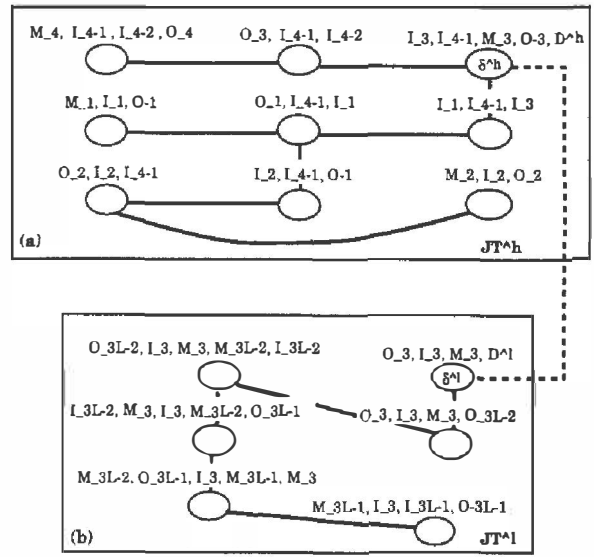

Figure 7: (a) $JT^h$ (b) $JT^l$. Adding the link shown as a dotted line creates the composite tree $JT^c$.

Now we construct a composite join tree $JT^c$ from $JT^h$ and $JT^l$. This is done by adding an link from $\delta^h$ to $\delta^l$ (shown as a dotted line in Fig 7). This composite join tree is a valid join tree for the network $B^c$ shown in Fig 5 (see next section for proof).

The composite join tree $JT^c$ has the following interesting property. If the user is not interested in details about the lower level nodes, then the update operation can be confined purely to the $JT^h$ segment of the join tree since only $JT^h$ has any variables of interest. More precisely, if there is no evidence available regarding the states of the lower level nodes *and* in addition, the user is not interested in details of the lower level nodes posterior distributions, then the update can be confined to $JT^h$.

Now suppose the user has finished an update in $JT^h$. She then decides that she does want to view more detail. In that case, the update process can be restarted and continued locally in $JT^l$. That is, the update process through the whole of $JT^h$ need not be repeated – the information coming from the rest of $JT^h$ is summarized in the message that $\delta_h$ sends $\delta_l$ when the update process is restarted. The restarted update process, in fact, is an incremental update which occurs only within $JT^l$. This incremental update can be performed at the user's demand – for example, in a graphical interface, the user may "open the window" corresponding to a "iconified" component. This can be interpreted as a request for more detailed information.

Along similar lines, if the user discovers evidence pertaining to a subcomponent, then she can "de-iconify" the containing component and assert the evidence. In this case, the update process begins in $JT^l$ and proceeds through $JT^h$ to make a global update. If one has multiple levels of hierarchy, the composite join tree has multiple levels of hierarchy too. At any time, the update process only affects that segment of the join tree that the user is interested in. This gives substantial savings in computation.

The dummy nodes $D^h$ and $D^l$ have been used only for ease of presentation. In practice, one only has to ensure that the join tree algorithm forces the nodes of interest to occur together in at least one clique.

### 3.3   $JT^c$ IS A VALID JOIN TREE

A valid join tree is constructed for a Bayesian network $B$ as follows [Pearl88]:
(1) The Bayesian network $B$ is converted into a Markov network $G$ by connecting the parents of each node in the network and dropping the directions of the arrows in the DAG. $G$ is an undirected graph.
(2) A chordal supergraph $G'$ is created from $G$ by a process called *triangulation*. A chordal graph is one where any loop of length 4 or more has a chord (an arc connecting two non-consecutive edges in the loop). Basically, the triangulation process adds arcs to the $G$ until it becomes chordal. (3) The maximal cliques of the chordal graph $G'$ are assembled into a tree $JT$. Each maximal clique is a vertex in the tree. The tree has the join tree property. The join tree property is the following: For every node $n$ of $B$, the sub-tree of $JT$ consisting purely of vertices which contain node $n$ is a connected tree.

It can be proved that $JT^c$ is a valid join tree for the Bayesian network $B^c$. We do so by first describing the construction of a particular chordal supergraph $G^{c'}$ of the Markov network of $B^c$. $JT^c$ is a valid join tree constructed from $G^{c'}$. We have included proof sketches below, the full proof is in [TechReport94].

Consider a graph $G^{c'}$ constructed as follows: $B^h$ is converted into a Markov network $G^h$. Similarly, $B^l$ is converted into a Markov network $G^l$. Each of these networks are triangulated giving the chordal graphs $G^{h'}$ and $G^{l'}$. $G^{h'}$ and $G^{l'}$ are merged to form a graph $G^{c'}$. This "merging" of the graphs is done as follows: The nodes $M_3$, $I_3$ and $O_3$ in $G^{h'}$ are merged with with the corresponding nodes in $G^{l'}$. That is, $G^{c'}$ has only one copy of each of these nodes. Any link between any of these nodes and a node in $G^{h'}$ is also present in $G^{c'}$. Similarly any link between any of these nodes and a node in $G^{l'}$ is also present in $G^{c'}$.

**Lemma 1**: $G^{c'}$ is a chordal supergraph of a Markov network $G^c$ of $B^c$.
**Proof sketch**: We note that the nodes in the set $S = \{M_3, I_3, O_3\}$ are the only nodes common to the subgraphs $G^{h'}$ and $G^{l'}$. Any loop $L$ that lies partly in $G^{l'}$ and $G^{c'}$ has to necessarily pass through $S$ twice. We see that in the $M_3$, $I_3$ and $O_3$ are necessarily connected to each other in both $G^{h'}$ and $G^{l'}$. Hence the loop $L$ has a chord that breaks it into two subloops $L^h$ and $L^l$ which lie purely in the chordal graphs $G^{h'}$ and $G^{l'}$ respectively. Hence $G^{c'}$ is chordal. It is easily proved that $G^{c'}$ is a supergraph of a Markov network $G^c$ of $B^c$. □

**Lemma 2**: $JT^c$ is a valid join tree created from $G^{c'}$.
**Proof sketch**: We note that any maximal clique in $G^{l'}$ which contains at least one node $n$ which does not occur in $G^{h'}$ is also a maximal clique in $G^{c'}$. We now observe that *every* maximal clique in $G^{l'}$ contains at least one node which does not occur in $G^{h'}$. We make a similar argument for the maximal cliques of $G^{h'}$. This implies that vertices of $JT^c$ are the maximal cliques of $G^{c'}$. We note that the running intersection property (r.i.p) holds for any node $n$ of $B^c$ which appears solely in $B^h$ (similarly, $B^l$) since $n$ appears purely in the vertices of true in $JT^h$ (similarly, $JT^l$). The only nodes which appear in both $B^h$ and $B^l$ are $M_3$, $I_3$ and $O_3$. Since these nodes appear both in $\delta^h$ and $\delta^l$ we see that the running intersection property holds for them too. □

**Theorem**: $JT^c$ is a valid join tree for the Bayesian network $B^c$.
**Proof**: This follows directly from Lemmas 1 and 2. □

The dummy nodes $D^h$ and $D^l$ are present solely to force a particular topology on the join trees $JT^h$ and $JT^l$. After the triangulation process they can be dropped from the cliques which contain them. This might sometime result in a simplification of the composite join tree. Consider the case where $\delta^l$ is reduced to $\{M_3, I_3, O_3\}$ after $D^l$ is dropped. In this situation, $\delta^l$ can be merged with $\delta^h$ since it is a subset of $\delta^h$. Similarly $\delta^h$ can be merged with $\delta^l$ if $\delta^l$ reduces to $\{M_3, I_3, O_3\}$ after $\delta^h$ is dropped. $JT^c$ continues to be a valid join tree after such mergers.

## 4  RELATED WORK

Geffner and Pearl [Geffner87] describe a scheme for doing distributed diagnosis of systems with multiple faults. They devise a message passing scheme by which, given an observation, a most likely explanation is devised. An explanation is an assignment of a mode state to every component in the schematic. The translation scheme described in this paper can be used to achieve an isomorphic result. That is, instead of using a Bayesian network update algorithm to compute updated probabilities of individual faults we could use a *dual* algorithm for computing composite belief [Pearl87] and compute exactly the same result. From the perspective of this paper, [Geffner87] have integrated the inference in the Bayesian network into the schematic as a message passing scheme. Separating out the network translation explicitly allows features such as hierarchical diagnosis, computation of updated probabilities in individual components as against component beliefs and many others (see below).

Mozetič [Mozetič92] lays out a formal basis for diagnostic hierarchies and demonstrates a diagnostic algorithm which takes advantage of the hierarchy. The approach is not probabilistic. However, he includes a notion of non-determinism in the following sense: Given the mode of a component he allows the input-output mapping of a component to be relation instead of a function – there can be multiple possible outputs for a given input. The notion of hierarchy we have described here corresponds to one of three possible schemes of hierarchical modeling that he describes. Our scheme can be expanded to support a probabilistic generalization of the other two schemes of modeling and his notion of non-determinism.

Genesereth [Genesereth84] describes a general approach to diagnosis including hierarchies. He distinguishes between structural abstraction and behavioral abstraction. In structural abstraction a component's function is modeled as the composition of the functions of subcomponents whose detail is suppressed at the higher level. This is similar to what we have described. Behavioral abstraction corresponds to a difference in how the function of a device is viewed – for example, in a low level description of a logic gate one might model input and output voltages while a high level description might model them as "high" and "low". Behavioral abstraction often corresponds to bunching sets of input values at the low level into single values at the higher level. Our method extends to support such abstractions in a straightforward manner.

Yuan [Yuan93] describes a framework for constructing decision models for hierarchical diagnosis. The decision model is comprised of the current state of knowledge, decisions to test or replace devices and a utility function that is constructed on the fly. A two step cycle comprising model evaluation and progressive refinement is proposed. The cycle ends when the fault is located (a single fault assumption is made). Model refinement is in accordance with the structural hierarchy of the device. The goal is to provide decision theoretic control of search in the space of candidate diagnoses. Such a framework needs a scheme for computing the relative plausibility of candidate diagnoses. Our work provides such a scheme in a general multiple fault setting.

## 5 CONCLUSION

The translation scheme described in this paper is a first step in an integrated approach to diagnosis, reliability engineering, test generation and optimal repair in hierarchically modeled dynamic discrete systems. The approach is probabilistic/utility-theoretic. We have made variety of assumptions in this paper for simplicity of exposition. The assumptions are: (a) non-correlated faults (b) full fault models (c) fully specified input distributions (d) components with single outputs (e) restricted form of hierarchy and (f) systems without dynamics or feedback. Each of these are relaxed in the general approach [Srinivas94]. We also discuss the temporal aspect of the "prior probability of failure" notion and relate it to standard quantities found in the reliability literature.